\documentclass[a4paper,twoside]{article}

\usepackage{epsfig}
\usepackage{subcaption}
\usepackage{calc}
\usepackage{amssymb}
\usepackage{amstext}
\usepackage{amsmath}
\usepackage{amsthm}
\usepackage{multicol}
\usepackage{pslatex}
\usepackage{apalike}
\usepackage[bottom]{footmisc}

\usepackage{csquotes} 
\usepackage{textcomp} 
\usepackage{gensymb} 
\usepackage{mathtools} 
\usepackage[shortlabels]{enumitem}

\usepackage{caption}
\graphicspath{ {images/} }

\usepackage[labelformat=simple]{subcaption}

\usepackage{multirow} 

\usepackage{SCITEPRESS}     

\begin{document}

\title{Multi-Phase Relaxation Labeling for Square Jigsaw Puzzle Solving}

\author{\authorname{Ben Vardi\sup{1}\orcidAuthor{0000-0002-6950-8297}, Alessandro Torcinovich\sup{2}\orcidAuthor{0000-0001-8110-1791}, Marina Khoroshiltseva\sup{2,3}\orcidAuthor{0000-0003-0424-0661}, Marcello Pelillo\sup{2,4}\orcidAuthor{0000-0001-8992-9243}\\and Ohad Ben-Shahar\sup{1}\orcidAuthor{0000-0001-5346-152X}}
\affiliation{\sup{1}Ben-Gurion University of the Negev, Be'er-Sheva, Israel}
\affiliation{\sup{2}Ca' Foscari University of Venice, Venice, Italy}
\affiliation{\sup{3}Istituto Italiano di Tecnologia, Genoa, Italy}
\affiliation{\sup{4}European Centre of Living Technology, Venice, Italy}
\email{\{benva@post, ben-shahar@cs\}.bgu.ac.il, \{ale.torcinovich, m.khoroshiltseva, pelillo\}@unive.it}
}


\abstract{We present a novel method for solving square jigsaw puzzles based on global optimization. The method is fully automatic, assumes no prior information, and can handle puzzles with known or unknown piece orientation. At the core of the optimization process is nonlinear relaxation labeling, a well-founded approach for deducing global solutions from local constraints, but unlike the classical scheme here we propose a multi-phase approach that guarantees convergence to feasible puzzle solutions. Next to the algorithmic novelty, we also present a new compatibility function for the quantification of the affinity between adjacent puzzle pieces. Competitive results and the advantage of the multi-phase approach are demonstrated on standard datasets.}

\onecolumn \maketitle \normalsize \setcounter{footnote}{0} \vfill

\section{\uppercase{Introduction}}
\label{sec:introduction}

The jigsaw puzzle game is a well-known and time-honored pastime for children and adults. Given $n$ non-overlapping image fragments, commonly referred to as puzzle pieces, the goal is to assemble them into a coherent visual image, preferably reconstructing the original (possibly unknown) image itself. Although large jigsaw puzzles are successfully solved by humans, the problem posed by this popular game is rather difficult, as it was shown to be NP-complete~\cite{demaine2007jigsaw}.
Algorithmic capability to solve jigsaw puzzles has important applications in several fields, such as information security~(Zhao, Su, Chou and Lee, 2007), assembly of shredded documents and photos~(Deever and Gallagher, 2012; Liu, Cao and Yan, 2011) and archaeology (Koller and Levoy, 2006).

\nocite{deever2012semi,zhao2007puzzle,liu2011automated,koller2006computer}

A particular type of jigsaw puzzles that has raised interest in the scientific community deals with \textit{square jigsaw puzzles} (e.g., Cho, Avidan and Freeman, 2010; Pomeranz, Shemesh and Ben-Shahar, 2011), where all pieces have identical square shape, geometric information is unavailable, and only pictorial information may be used for driving the solution. This is also why solving square jigsaw puzzles marks an upper bound of jigsaw puzzle reconstruction performance when geometric cues are diminishing. Such an understanding is a key step in developing solvers for real-world puzzles, in which both pictorial and geometric information are usually available in varying degrees.

\nocite{cho2010probabilistic,pomeranz2011fully}

Square jigsaw puzzle solvers may be distinguished by two criteria. The first is the evaluation method for potential piece matchings, which defines the compatibility of placing any two puzzle pieces in one of the possible spatial 4-neighborhood adjacency relations. The measure used for this evaluation is commonly referred to as \textit{piece compatibility measure}. The second criterion is the algorithmic method used to process these compatibilities in order to infer a solution for the full puzzle. In other words, one may argue that square jigsaw puzzle solvers use local constraints (piece compatibilities) in order to deduce a global solution (piece placements that seek to restore the original image).

While several attempts were made to solve square jigsaw puzzles and impressive results were achieved (e.g., Pomeranz et~al., 2011; Andal{\'o}, Taubin and Goldenstein, 2016; Son, Hays and Cooper, 2018), most of the suggested algorithms use greedy methods rather than global optimization methods. Nevertheless, global optimization algorithms are generally preferred over greedy algorithms for their ability to produce solutions which simultaneously take into consideration all the local potential piece matchings. Moreover, global optimization approaches may often facilitate additional mathematical analysis~\cite{floudas2013deterministic} and thus a better understanding of the dynamics of the problem and solution process.

\nocite{andalo2016psqp,son2018solving}

A well-founded global optimization approach that handles problems where local constraints are given and global interpretation should be found is \textit{relaxation labeling} (RL). The approach, first introduced by Rosenfeld, Hummel and Zucker~(1976), is a class of iterative procedures for refining label assignments based on contextual constraints. RL has been applicable in solving many global optimization problems, most commonly from the computer vision field~(e.g., Zucker, Hummel and Rosenfeld, 1977). Although iterative, and reminiscent of gradient ascent, it can be formulated without any step size~\cite{pelillo1997dynamics,rosenfeld1976scene} while guaranteeing the optimization of its objective function~\cite{hummel1983foundations,pelillo1997dynamics}.

\nocite{rosenfeld1976scene,hummel1983foundations,zucker1977application}

Recently, preliminary results of using RL for square jigsaw puzzle solving~\cite{khoroshiltseva2021jigsaw} demonstrated the potential of this approach, but nevertheless suffered from several limitations such as occasional failures to converge to a feasible solution or being limited to puzzles in which piece orientations are known (known as Type 1 puzzles; Gallagher, 2012).
In this work we further explore the potential of RL for square jigsaw puzzle solving and present a new RL algorithm that does guarantee feasible solutions, performs better than the preliminary approach, and handles both Type 1 but also Type 2 puzzles (whose piece orientations are \textit{unknown}).
In addition to the algorithmic novelty, another contribution of this paper is the proposal of a new piece compatibility measure.

\nocite{gallagher2012jigsaw}

\section{\uppercase{Related Work}}

\subsection{Puzzle Solving}

The topic of puzzle solving has long been of interest in the scientific community. The first to investigate puzzles were Freeman and Garder~(1964), focusing on the geometric information of irregularly shaped puzzle pieces to suggest the first computational \textit{apictorial} solver. Three decades later, Kosiba, Devaux, Balasubramanian, Gandhi and Kasturi~(1994) were the first to also consider \textit{pictorial} puzzles in their classical \enquote{toy} jigsaw puzzle solver.

\nocite{freeman1964apictorial,kosiba1994automatic}

While the effort to combine pictorial and geometric information was followed also by others, the line of research that has been dominating the literature focused on square jigsaw puzzles in which all pieces have an identical square shape, with puzzle dimensions being often known. While the objective of all different square jigsaw puzzle solvers is the same, solution approaches vary greatly.
A first notable attempt was made by Cho et~al.~(2010) who employed loopy belief propagation algorithm for Type 1 puzzles. Although being able to solve the biggest puzzle at the time, their algorithm was only semi-automatic, requiring at least some ground truth information. A year later, Pomeranz et~al.~(2011) published the first fully automatic solver, requiring no prior knowledge. Their iterative greedy Type 1 solver is based on three steps: pieces placement according to the piece compatibility measure, segmentation of the proposed solution into coherently assembled segments, and shifting segments to improve the current solution. Despite being simpler, this approach provided not only a leap in performance, but also the ability to solve puzzles an order of magnitude larger than before.

Following Pomeranz et~al.~(2011), Andal{\'o} et~al.~(2012; 2016) were the first to formulate the problem as a global optimization problem using a constrained quadratic objective function. 
Gallagher~(2012) used a tree-based greedy algorithm inspired by the well-known Kruskal's algorithm for finding minimum spanning trees. Gallagher~(2012) was the first to introduce and solve Type 2 puzzles, namely puzzles in which piece orientations are unknown. 
Shortly after, Sholomon, David and Netanyahu~(2013) devised a genetic algorithm-based solver for Type 1 puzzles, and later extended it for Type 2 puzzles (Sholomon, David and Netanyahu, 2014).

A different approach for the problem was proposed by Son et~al.~(2014; 2018) who devised a Type 1 and 2 solver that utilizes loops of pieces that form consistent cycles. More recently, Huroyan, Lerman and Wu~(2020) proposed a Type 2 method based on recovering piece orientations by estimating the graph connection Laplacian.
Of special importance to our work is the global optimization solver by Khoroshiltseva et~al.~(2021), which is based on RL. Due to its relevance, we dedicate Section~\ref{subsec:related_work:puzzle_solving_with_balancing} to that solver.

\nocite{andalo2012solving,sholomon2013genetic,sholomon2014generalized,khoroshiltseva2021jigsaw, huroyan2020solving, son2014solving}

Several neural network methods have been suggested to solve square jigsaw puzzles, most of them applicable only for small puzzles.
Paumard, Picard and Tabia~(2020) offered a convolutional neural network-based solver that predicts neighboring relations in 9 pieces puzzles. Li, Liu, Wang, Liu and Zeng~(2021) devised a generative adversarial network (GAN) pipeline for training a classification network to predict pieces permutation in puzzles with up to 16 pieces. Talon, Del Bue and James~(2022) suggested another GAN-based approach that is capable of solving bigger puzzles, but suffers from diminishing accuracy as puzzle size gets bigger.

\nocite{paumard2020deepzzle, li2021jigsawgan, talon2022ganzzle}

Along with the different algorithmic approaches to the square puzzle problem, different piece compatibility measures have been also developed over time. Such measures define the compatibility of placing any two puzzle pieces in any of the 4-neighborhood adjacency relations. They are most commonly defined as a function $C(b_i, b_j, R)$ that specifies how compatible it is to place piece $b_j$ in relation $R \in \{right, down, left, up\}$ to piece $b_i$. In virtually all solvers, piece compatibilities are based on a more fundamental measure of piece dissimilarity that usually compares the pictorial information along the boundaries of pieces. This measure is later mapped to a score of compatibility.

\subsection{Relaxation Labeling}

Relaxation labeling (RL) is a class of processes for ambiguity reduction of labeling assignments, done by iteratively refining assignments based on contextual information~\cite{rosenfeld1976scene}. Among these processes, in the current work we focus on \textit{nonlinear relaxation labeling}, where problems involve four elements: 

\begin{enumerate}[(1), noitemsep]
    \item Set of $n$ objects $B=\{b_1,...,b_n\}$.
    \item Set of $m$ possible labels $\Lambda = \{\lambda_1,...,\lambda_m\}$.
    \item Four-dimensional matrix of \textit{compatibility coefficients} ${\{r_{ij}(\lambda, \mu)\}} \in \mathbb{R}^{n^{2}m^{2}}$, expressing how compatible are the two hypotheses of \enquote{labeling object ${b_i}$ with label ${\lambda}$} and \enquote{labeling object ${b_j}$ with label ${\mu}$}.
    \item \textit{Initial labeling} ${\Bar{p}^{(0)} \in \mathbb{R}^{nm}}$, containing prior probabilities for labeling each object ${b_i}$ with each label ${\lambda}$.
\end{enumerate}

The purpose of nonlinear relaxation labeling (henceforth referred to as just RL) is to provide a consistent labeling assignment ${\Bar{p} \in \mathbb{R}^{nm}}$ that represents a viable solution to the problem. An assignment is a set of probabilities for assigning each label $\lambda$ to each object $b_i$. While the initial labeling ${\Bar{p}^{(0)}}$ is already qualified as such, it may not be \enquote{good enough} in terms of satisfying the local constraints represented by the compatibility coefficients. An assignment that does satisfy these constraints is called \textit{consistent}, and RL is the optimization process that provides such consistent assignments given the four elements above.

A labeling assignment ${\Bar{p}}$, usually referred to just as labeling, is represented in the current work as \textit{labeling matrix} of $n$ rows and $m$ columns. The probability of object ${b_i}$ to be labeled with label ${\lambda}$ is notated as ${p_{i}(\lambda)}$, and is indicated by the matrix entry in row $i$ and column $\lambda$. 
Since each row $i$ represents the probabilities of all label assignments to object ${b_i}$, the space of all possible labelings is
\begin{equation}\label{eq:k_all_possible_labelings}
\small
\begin{split}
    \mathbb{K} \; = \; \bigg\{ \Bar{p} \in \mathbb{R}^{nm} \; | & \; p_{i}(\lambda) \geq 0, \; \forall i \in B, \lambda \in \Lambda
    \\ &
    \text{and } \sum_{\lambda = 1}^{m} p_{i}(\lambda) = 1, \; \forall i \in B \bigg\}\,.
\end{split}
\end{equation}
A particular important subset of $\mathbb{K}$ is the set of all possible \textit{unambiguous labelings}:
\begin{equation}\label{eq:k_star_all_binary_labelings}
\small
    \mathbb{K}^{*} \; = \; \bigg\{ \Bar{p} \in \mathbb{K} \; | \; p_{i}(\lambda) \in \{0,1\}, \; \forall i \in B, \lambda \in \Lambda \bigg\}\,,
\end{equation}
where each object is assigned one label with full confidence.

In order to refine the initial (likely inconsistent) labeling, RL algorithm involves the computation of two terms in each iteration $k$ of the process, the \textit{support} and a new more \enquote{refined} labeling. While not necessarily the only way to do so, we adapt the common terms from the RL literature~\cite{pelillo1997dynamics,rosenfeld1976scene}:

\begin{enumerate}[(1), noitemsep]
    \item The support matrix ${\Bar{q}^{(k)} \in \mathbb{R}^{nm}}$ represents the contextual \enquote{support} that each object $b_i$ gets for labeling it with each label $\lambda$. This matrix, referred to as \textit{support}, is computed by a \textit{support function}:
    \begin{equation}\label{eq:rl_support}
        \small
        q_i^{(k)}(\lambda) = \sum_{j=1}^{n} \sum_{\mu=1}^{m} r_{ij}(\lambda, \mu)p_j^{(k)}(\mu)\,.
    \end{equation}
    
    \item The refined labeling $\Bar{p}^{(k+1)}$ that represents the new \enquote{better} labeling, is computed by a \textit{nonlinear update rule}:
        \begin{equation}\label{eq:rl_update_rule}
        \small
        p_i^{(k+1)}(\lambda) = \frac{p_i^{(k)}(\lambda)q_i^{(k)}(\lambda)}{\sum_{\mu=1}^{m} p_i^{(k)}(\mu)q_i^{(k)}(\mu)}\,.
    \end{equation}
\end{enumerate}

The entire algorithm is best viewed as a continuous mapping of $\mathbb{K}$ onto itself through Eq.~\ref{eq:rl_update_rule}, and it is expected to converge to a fixed point if applied successively. Theoretically, such a fixed point iteration stops when $p_i^{(k+1)}(\lambda) = p_i^{(k)}(\lambda), \; \forall i, \lambda$ for some iteration $k$. In practice, we may (heuristically) determine convergence if the increase of the global objective function (see below in Eq.~\ref{eq:rl_alc}) in two successive iterations is smaller than a given threshold $\epsilon$.

The theoretical foundations for RL were laid by Hummel and Zucker~(1983) who have proved several useful theoretical properties for an alternative RL process. Pelillo~(1997) later proved similar dynamics for the original RL algorithm from Rosenfeld et~al.~(1976). Of particular relevance to us is that under symmetric non-negative compatibility coefficients (i.e., $r_{ij}(\lambda, \mu) = r_{ji}(\mu, \lambda) \geq 0,\; \forall i,j,\lambda,\mu$), each iteration of the process strictly increases a global consistency measure known as \textit{average local consistency} (ALC; Hummel and Zucker, 1983):

\begin{equation}\label{eq:rl_alc}
\small
A(\bar{p}, \bar{q}) = \sum_{i=1}^{n} \sum_{\lambda=1}^{m} p_i(\lambda)q_i(\lambda)\,.
\end{equation}
Both contributions~\cite{hummel1983foundations,pelillo1997dynamics} showed that their respective process eventually approaches and converges to a consistent solution near the initial labeling assignment. Importantly, however, unlike other optimization algorithms such as gradient ascent, the original iteration due to Rosenfeld et~al.~(1976) does so without the need to determine a step size.

\subsection{Puzzle Solving with ``Balanced'' RL}
\label{subsec:related_work:puzzle_solving_with_balancing}

To our best knowledge, the only work that employs RL for square jigsaw puzzle solving is by Khoroshiltseva et~al.~(2021). In their formulation for Type 1 puzzles, the set of objects is the set of puzzle pieces and the set of labels is the set of possible positions on the puzzle grid. In these settings $n = m$, so the space $\mathbb{K}$ is the space of stochastic matrices, and the space $\mathbb{K}^{*}$ is the space of binary stochastic matrices. As no better prior information can be assumed, the initial labeling is a uniform distribution of labels across each object. The compatibility coefficients $r_{ij}(\lambda, \mu)$ are set to piece compatibility scores $C(b_i, b_j, R)$ for all pairs of different pieces $b_i, b_j$ and positions $\lambda, \mu$ that represent neighboring positions on the puzzle grid in relation $R$. In all other cases, $r_{ij}(\lambda, \mu)$ is $0$. 

Unfortunately, applying the RL update rule from Eq.~\ref{eq:rl_update_rule} using such definitions does not guarantee, and in fact often does not result in a \textit{feasible} puzzle solution, defined as an assignment where each piece is placed at a unique position on the puzzle grid and all positions are occupied with some piece. Such feasible solutions, represented by \textit{permutation labeling matrices}, are not always achieved because the theory just guarantees that the update rule will map a stochastic matrix into another stochastic matrix, with the additional hope that the process converges to an unambiguous assignment, namely a binary stochastic matrix. But even if the process converges to a binary stochastic matrix (as opposed to a non-binary one), this does not prohibit situations where two objects (pieces) obtain the same label (position), or that a given label (position) is not assigned to any object (piece).
To remedy this situation, Khoroshiltseva et~al.~(2021) extended the update rule from Rosenfeld et~al.~(1976) with two versions of a matrix balancing component that seeks to map a stochastic labeling matrix to a \textit{doubly stochastic} matrix. Since the latter are easier to binarize into permutation matrices, this should help the process converge to the space of all feasible puzzle solutions.

Khoroshiltseva et~al.~(2021) showed empirically that their approach can perform better than a plain RL without balancing. At the same time, while the method encourages feasible solutions, it cannot guarantee them. Testing on natural images indeed showed vulnerability to certain common conditions, and in particular the presence of constant pieces, namely puzzle pieces of identical appearance of some constant value throughout. Such failure cases can be explained by the fact that the rows of the labeling matrix that refer to constant pieces are identical in all steps of the process (both the update rule and balancing component map identical rows into identical rows), thus the process cannot converge to a permutation matrix.

In the current work we suggest an alternative and improved puzzle solver based on a novel multi-phase RL scheme. Our method is applicable for both Type 1 and 2 puzzles, addresses the aforementioned problems and facilitates further exploration of RL as a computational powerhouse for puzzle solving.

\section{\uppercase{Multi-phase RL Puzzle Solvers}} 

\subsection{Puzzles with Known Piece Orientation}

Type 1 puzzles, possibly the most elementary variation of the puzzle problem, constitute unordered set of square pieces whose orientation is given but their organization in an $N \times M$ array of positions is not. Treating this problem as an RL problem can be done in the following way, similar to Khoroshiltseva et~al.~(2021):

\begin{enumerate}[(1), noitemsep]
    \item The objects $B=\{b_1,\ldots,b_n\}$ are puzzle pieces.

    \item The labels $\Lambda = \{\lambda_1,...,\lambda_m\}$ are all possible positions $(x_\lambda, y_\lambda)$ on the two-dimensional $N \times M$ puzzle grid.

    \item The compatibility coefficients are defined by
    \begin{equation}\label{eq:rl_comp_coeffs}
    \small
    \begin{split}
    & r_{ij}(\lambda, \mu) =  r_{ij}((x_\lambda,y_\lambda), (x_\mu,y_\mu))=\\
        & \begin{cases*}
            C(b_i, b_j, R), & if $i \neq j$ and $(x_\lambda,y_\lambda), (x_\mu,y_\mu)$ are \\
            & neighboring positions in relation $R$,\\
            0, & otherwise\,,
    \end{cases*}
    \end{split}
    \end{equation}
    where $R \in \{right, down, left, up\}$ represents a 4-neighborhood spatial relation and $C$ is a non-negative and symmetric piece compatibility function (see Section~\ref{sec:piece_compatibility} for the function definition).
  
    \item The initial labeling $\bar{p}^{(0)}$ is the barycenter of the search space, i.e., $p_i^{(0)}(\lambda) = \frac{1}{m},\; \forall i, \lambda$. With no prior information, this is the only rational choice one can make. 
\end{enumerate}

\begin{figure}
\centering
\begin{subfigure}[b]{0.15\textwidth}
    \centering
    \includegraphics[height=20mm]{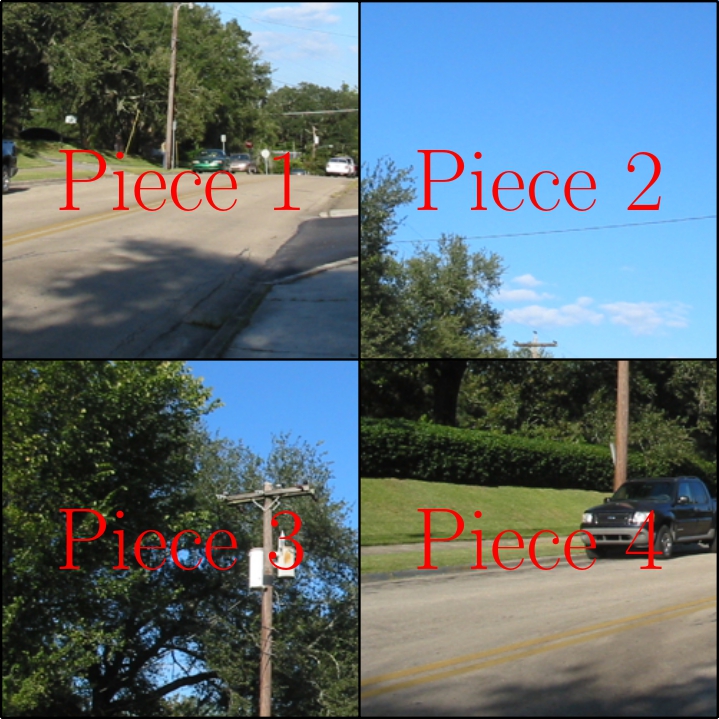}
    \caption{}
    \label{fig:type_1_labeling:puzzle}
\end{subfigure}
\begin{subfigure}[b]{0.15\textwidth}
    \centering
    \includegraphics[height=20mm]{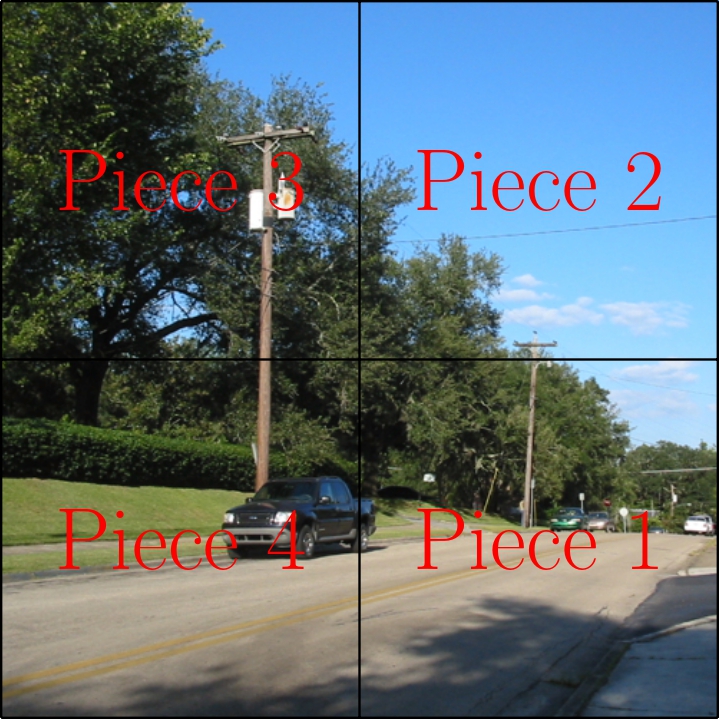}
    \caption{}
    \label{fig:type_1_labeling:solution}
\end{subfigure}
\begin{subfigure}[b]{0.15\textwidth}
    \centering
    \includegraphics[height=21mm]{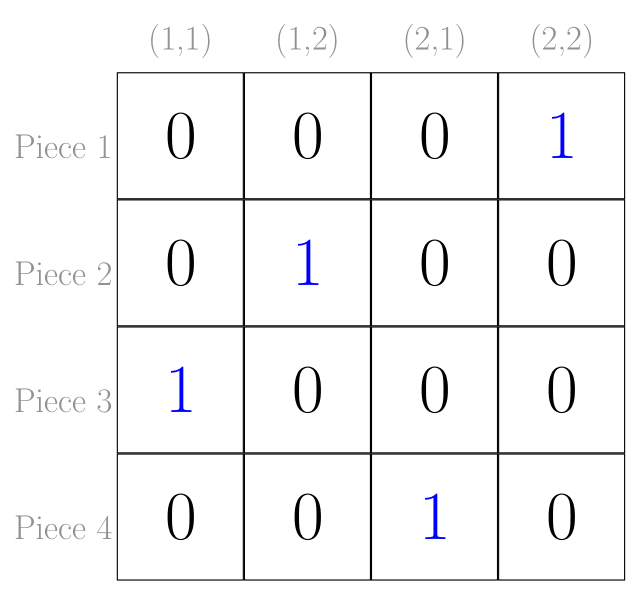}
    \caption{}
    \label{fig:type_1_labeling:labeling}
\end{subfigure}
\caption{Type 1 puzzle feasible labeling structure. \subref*{fig:type_1_labeling:puzzle} is a $2 \times 2$ puzzle and \subref*{fig:type_1_labeling:solution} is its  perfect solution. This perfect solution is represented by the permutation labeling matrix \subref*{fig:type_1_labeling:labeling}. Piece IDs in red were added for demonstration purposes.}
\label{fig:type_1_labeling}
\end{figure}

As mentioned above, a feasible Type 1 reconstruction is a solution in which each piece is placed at a unique position on the puzzle grid, and the space of such feasible labelings is the space of permutation matrices. Figure~\ref{fig:type_1_labeling} exemplifies feasible labeling structure for the Type 1 problem.

Since $C$ is non-negative and symmetric, the same applies to $r_{ij}(\lambda, \mu)$. Consequently, the process is guaranteed to strictly increase the ALC in each iteration, and to converge to a consistent assignment~\cite{pelillo1997dynamics}.
For puzzle solving this turns the ALC to an intuitive objective function and the relaxation process to one that seeks to maximize the sum of compatibilities between all adjacent pieces, similar in spirit to other global optimizers for puzzle solving~\cite{andalo2016psqp,sholomon2013genetic}. For the justification of this claim please refer to the supplementary material in the project webpage
(\textit{https://icvl.cs.bgu.ac.il/multi-phase-rl-for-jigsaw-puzzle-solving}).

\subsubsection{Type 1 Solver}

With the elements of the RL process now set for puzzle solving, and the observation that preliminary attempts~\cite{khoroshiltseva2021jigsaw} could not guarantee feasible solutions, we suggest an alternative algorithm that does guarantee this property. Central to our approach is a novel multi-phase scheme where each phase uses RL to determine the position of a \textit{single} puzzle piece at a \textit{non-occupied} position. As explained below, this enables to change the structure of the labeling assignment towards permutation labeling, as required for feasible solutions.

\begin{figure}
\centering
\begin{subfigure}[b]{0.23\textwidth}
    \centering
    \includegraphics[height=36.5mm]{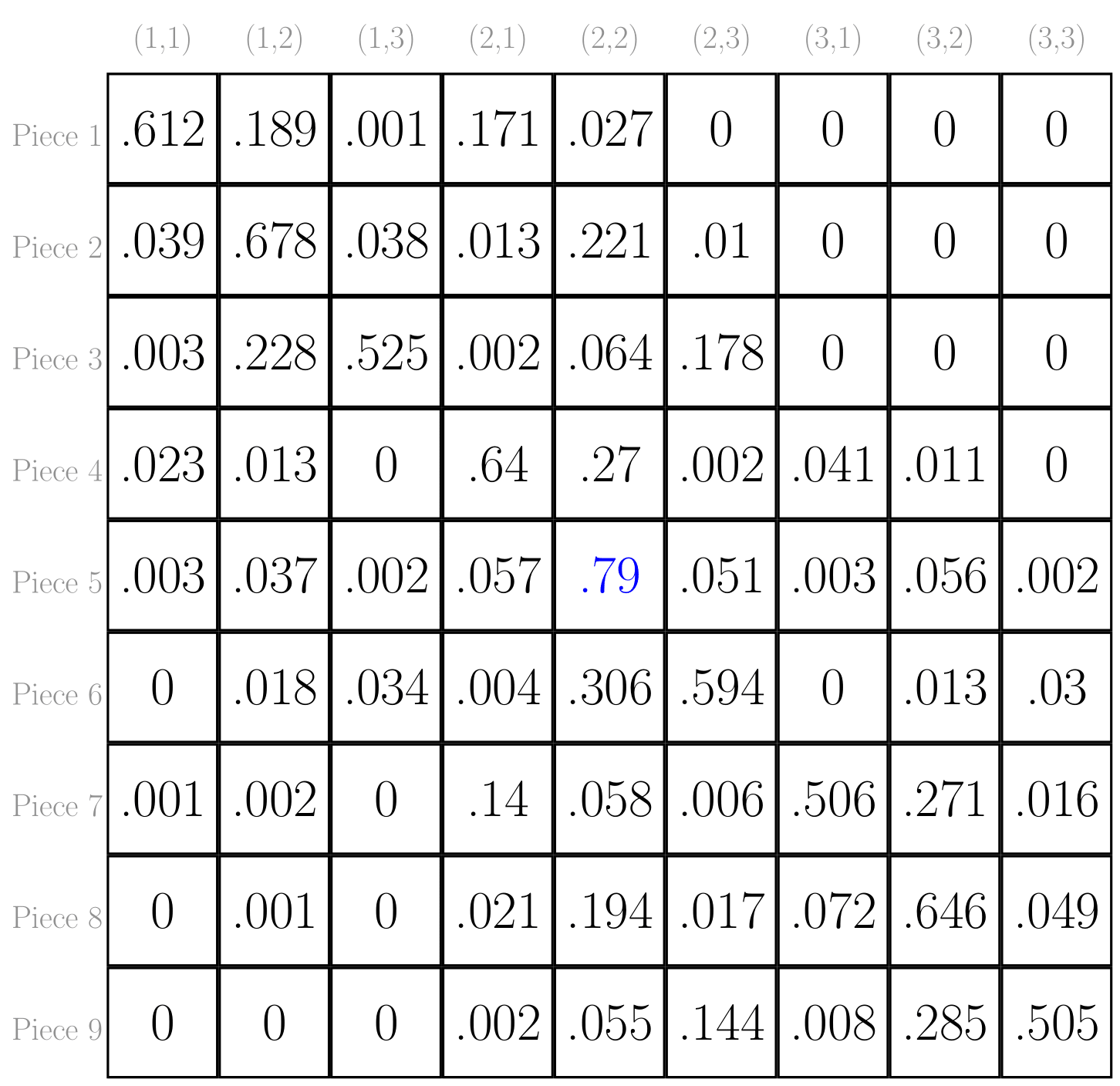}
\end{subfigure}
\hfill
\begin{subfigure}[b]{0.23\textwidth}
    \centering
    \includegraphics[height=36.5mm]{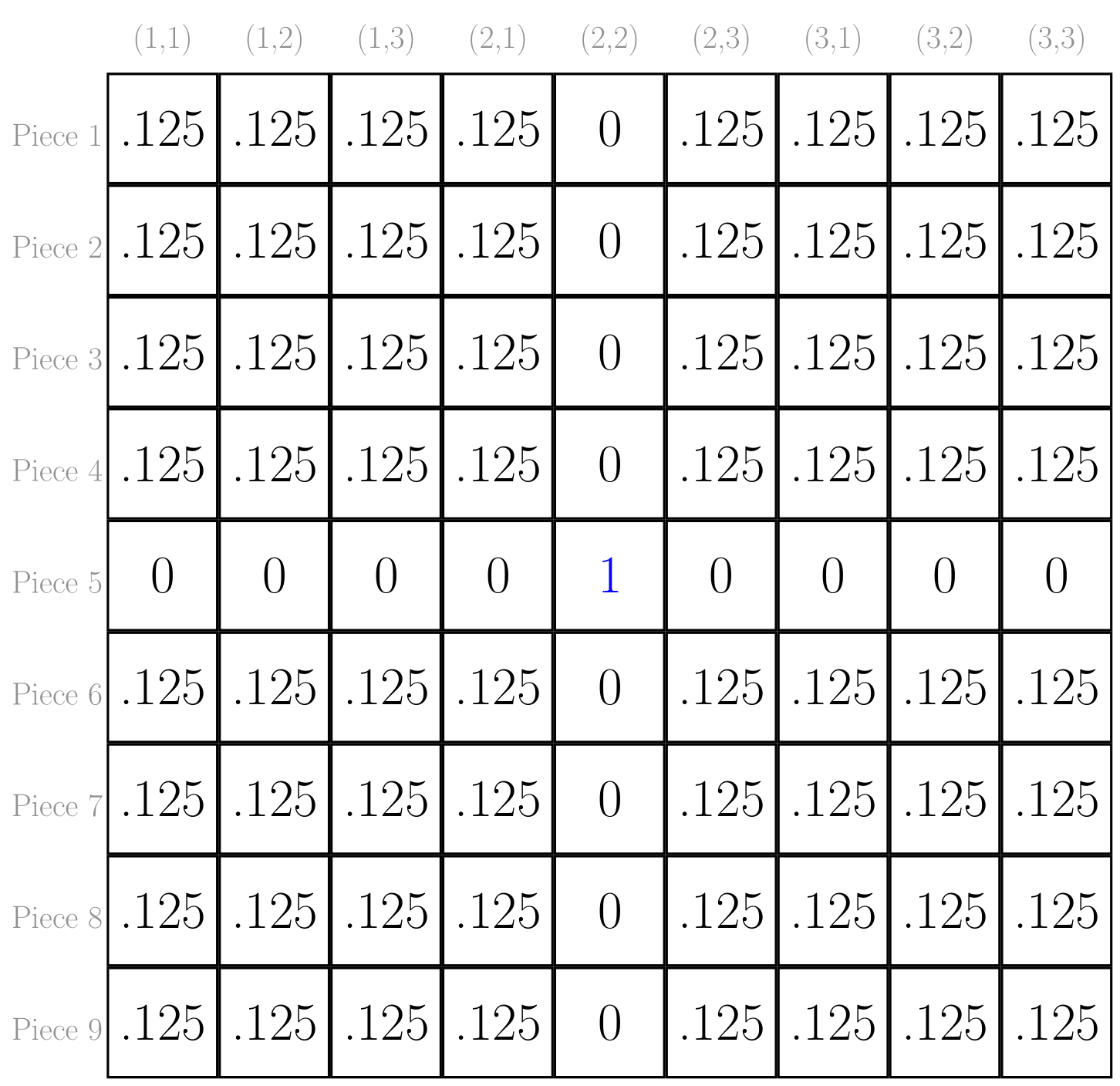}
\end{subfigure}
\caption{
The first phase of our Type 1 solver on a $3~\times~3$ puzzle, done with $\alpha = 0.7$. The RL phase begins from the initial labeling and terminates once some labeling value crosses $\alpha$ (left matrix, obtained for $p_5((2,2))$ after $4$ RL iterations). On the right is the labeling after anchoring piece $b_5$ to position $(2,2)$.}
\label{fig:type_1_solver_phase}
\end{figure}

More specifically, in each phase of the algorithm we let the standard RL algorithm~\cite{pelillo1997dynamics,rosenfeld1976scene} run and intervene only when the process is confident enough about placing a certain piece $b_i$ at some position $\lambda$, namely that the process converges to a consistent labeling and/or at least one label at one object satisfies $p_i(\lambda) \geq \alpha$, where $\alpha$ is a pre-defined threshold. 
This intervention marks the end of the current phase, and involves three steps: 
\begin{enumerate}[(1), noitemsep]
    \item \textit{Anchoring} piece $b_i$ to position $\lambda$ and disallowing placing it from now on at any other position. This is done by setting $p_i(\lambda) = 1$ and  $p_i(\mu)~=~0,\; \forall~\mu~\neq~\lambda$. 
    \item Preventing the future placement of other pieces at position $\lambda$. This is done by setting $p_j(\lambda)~=~0,\; \forall~j~\neq~i$.
    \item Resetting all entries $p_j(\mu)$ that involve pieces $b_j$ and positions $\mu$ for which past phases have not made a decision yet. This step is done by setting these entries to the barycenter of the remaining search \textit{subspace}, i.e.~to the reciprocal of the number of remaining unassigned objects/positions. This allows subsequent phases to start afresh without bias from prior phases.
\end{enumerate}
Figure~\ref{fig:type_1_solver_phase} demonstrates these steps after the completion of the first RL phase on a $3 \times 3$ puzzle.

The manipulated labeling serves as the initial labeling for the next phase of the algorithm, which anchors another piece at another position. The algorithm finally terminates after anchoring all pieces, namely after $n$ phases. Importantly, although no special measures are taken, the process does not modify entries of pieces and positions for which previous phases have already made a decision since Eq.~\ref{eq:rl_update_rule} does not (indeed, cannot) modify binary (i.e., 0 or 1) probabilities. 
Thanks to this property, each phase takes a decision about a single piece and position by binarizing their respective row and column in the labeling matrix, all while utilizing the information obtained in previous phases. This way, each phase updates the labeling so it becomes progressively ``closer'' to a permutation labeling. After $n$ phases convergence to a permutation labeling is guaranteed. 

Operating as described above, the multi-phase solver will normally attach new pieces to the block of pieces already anchored in the past. However, there is no guarantee for this behavior. To prevent the solver from generating isolated \enquote{islands} of solutions that would be difficult to \enquote{stitch} together properly across the puzzle grid, we allow anchoring only at positions adjacent to previously anchored ones, thus guaranteeing a single contiguous block of anchored pieces. 
Moreover, if the $\alpha$ threshold is met simultaneously by more than one combination of piece $b_i$ and position $\lambda$, we select a single combination by an internal sorting of relevant combinations. Please see the supplementary material for more details.

\begin{figure}
\centering
\begin{subfigure}[b]{0.11\textwidth}
    \centering
    \includegraphics[height=13.1mm]{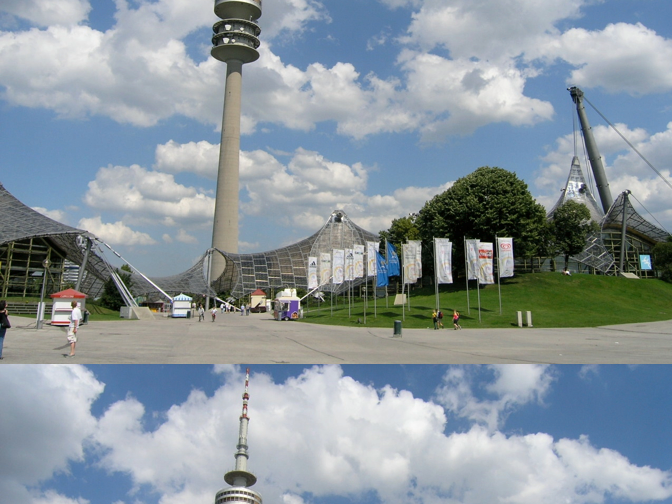}
    \caption{}
    \label{subfig:type_1_block_translation:shifted_solution}
\end{subfigure}
\begin{subfigure}[b]{0.11\textwidth}
    \centering
    \includegraphics[height=13.1mm]{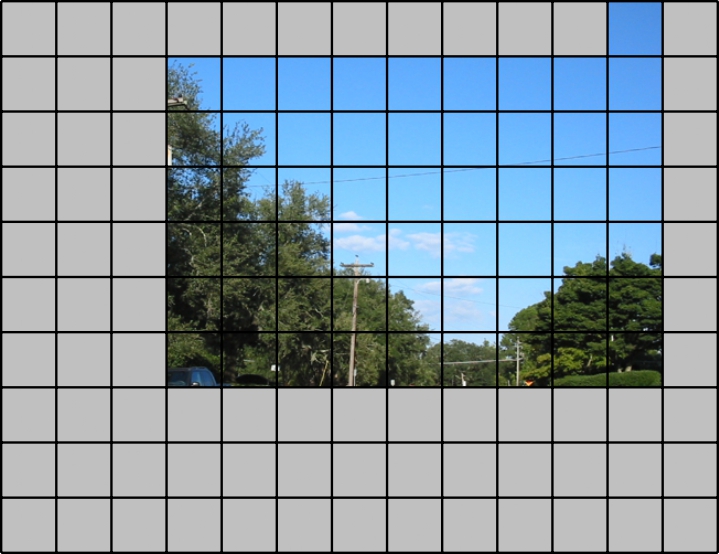}
    \caption{}
    \label{subfig:type_1_block_translation:after_phase_no_trans}
\end{subfigure}
\begin{subfigure}[b]{0.11\textwidth}
    \centering
    \includegraphics[height=13.1mm]{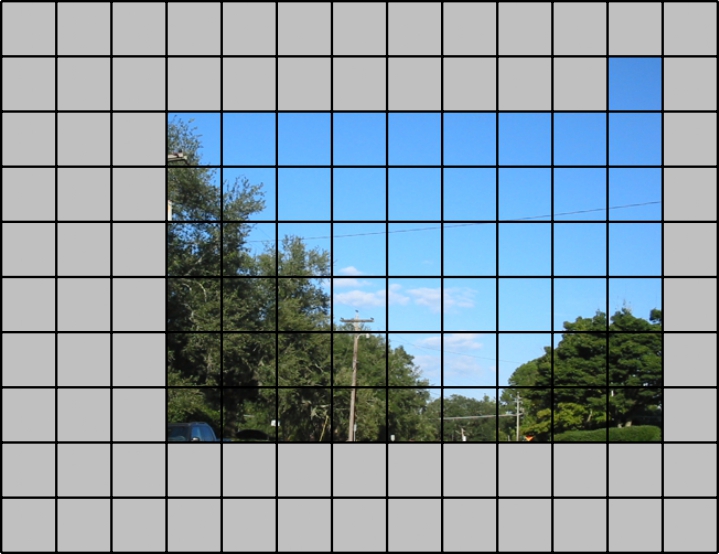}
    \caption{}
    \label{subfig:type_1_block_translation:after_phase_with_trans}
\end{subfigure}
\begin{subfigure}[b]{0.11\textwidth}
    \centering
    \includegraphics[height=13.1mm]{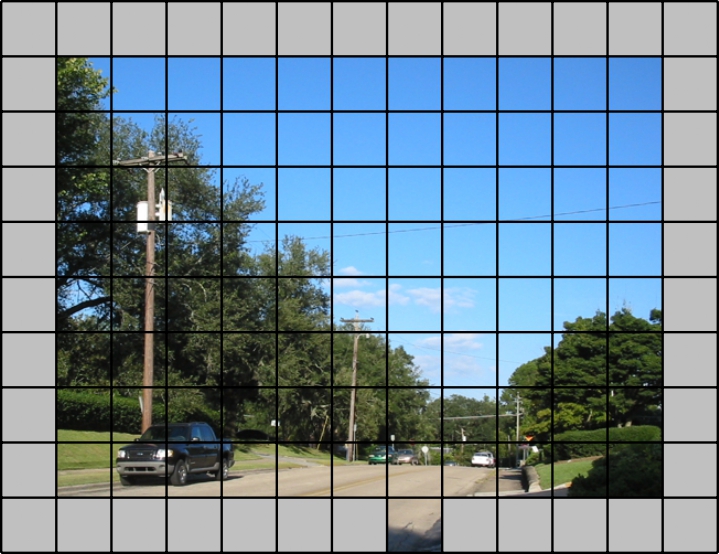}
    \caption{}
    \label{subfig:type_1_block_translation:dilemma}
\end{subfigure}
\caption{The shifted reconstruction phenomenon and its solution. \subref*{subfig:type_1_block_translation:shifted_solution} is a reconstruction where rows are shifted. To prevent this, we translate the whole anchored block when the anchoring of some piece reaches the edge of the puzzle grid, as in \subref*{subfig:type_1_block_translation:after_phase_no_trans} (gray areas represent non-occupied positions). At this point we translate the block downwards to obtain \subref*{subfig:type_1_block_translation:after_phase_with_trans} and allow subsequent phases to grow in all directions. \subref*{subfig:type_1_block_translation:dilemma} is a case a translation will not necessarily benefit the situation as translating the block upwards will also limit the growth. At this point we branch the computation to try both options, translating the block upwards or leaving the block as is.}
\label{fig:type_1_block_translation}
    
\end{figure}

The algorithm just described guarantees \textit{feasible} solutions although, of course, it cannot guarantee \textit{correct} solutions.
In practice, many solutions are perfect (or near perfect) up to a circular shift of the rows and/or columns, as shown in Figure~\ref{subfig:type_1_block_translation:shifted_solution}.
This phenomenon is best explained as a lack of awareness of the solver to the puzzle dimensions, as it emerges from the likely event of assigning the very first piece to a wrong position in the puzzle grid. This problem was also observed in past (non-RL) solvers (e.g., Pomeranz et~al., 2011), but is easy to alleviate by incorporating such awareness in the process. In particular, each time a phase is completed, if the anchored block abuts the edge of the puzzle grid, we translate it one column or row inside (see Figures~\ref{subfig:type_1_block_translation:after_phase_no_trans}-\ref{subfig:type_1_block_translation:after_phase_with_trans}). This translation is done by applying a proper transformation on the labeling matrix, and it allows subsequent phases to again grow the block in all directions. 

Of course, these translations of the anchored block will cease once the solution is short of one row or column to the vertical or horizontal dimension of the puzzle (e.g., Figure~\ref{subfig:type_1_block_translation:dilemma}). At this point it is yet impossible to know what side the coming phases should grow the last row or column and thus we branch the algorithm to try both options. Doing this for both the vertical and horizontal growth, we end up with 4 final reconstructions, from which the final solution is the one with the largest ALC.

\subsection{Puzzles with Unknown Piece Orientation}\label{sec:Type_2_Solver}

Type 2 puzzles extend the elementary case to situations where piece orientation is unknown. Since the final orientation of the pieces needs to be determined by the process, the setup of the problem as an RL process requires some adjustments. 
As before, the objects $b_1,\ldots,b_n$ still represent the $NM$ puzzle pieces. However, the labels $\lambda_1,...,\lambda_m$ now need to represent all combinations of piece positions and orientations. Each label $\lambda$ is a pair $((x_\lambda, y_\lambda), \theta_\lambda)$, where $(x_\lambda, y_\lambda)$ is a position on the $N \times M$ grid and $\theta_\lambda \in \{$0\degree, 90\degree, 180\degree, 270\degree$\}$ is one of 4 possible piece orientations. 

Clearly, the change to labels requires a corresponding adjustment to the compatibility coefficients:
\begin{equation}\label{eq:rl_comp_coeffs_type2}
\small
\begin{split}
& r_{ij}(\lambda, \mu) =  r_{ij}(((x_\lambda,y_\lambda),\theta_\lambda), ((x_\mu,y_\mu),\theta_\mu))=\\
    & \begin{cases*}
        C(\theta_\lambda(b_i), \theta_\mu(b_j), R), & if $i \neq j$ and $(x_\lambda,y_\lambda), (x_\mu,y_\mu)$\\
        & are neighboring positions\\
        & in relation $R$,\\
        0, & otherwise\,,
\end{cases*}
\end{split}
\end{equation}
where function $\theta_\lambda(b_i)$ represents the clockwise-rotation of piece $b_i$ by $\theta_\lambda$ degrees.

\begin{figure}
\centering
\begin{subfigure}[b]{0.23\textwidth}
    \centering
    \includegraphics[width=20mm]{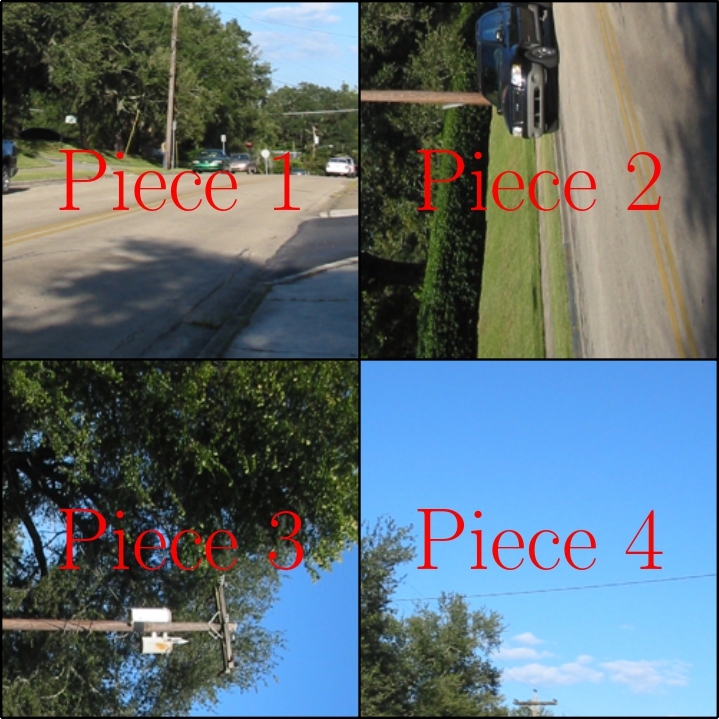}
    \caption{}
    \label{subfig:type_2_labeling:puzzle}
\end{subfigure}
\begin{subfigure}[b]{0.23\textwidth}
    \centering
    \includegraphics[width=20mm]{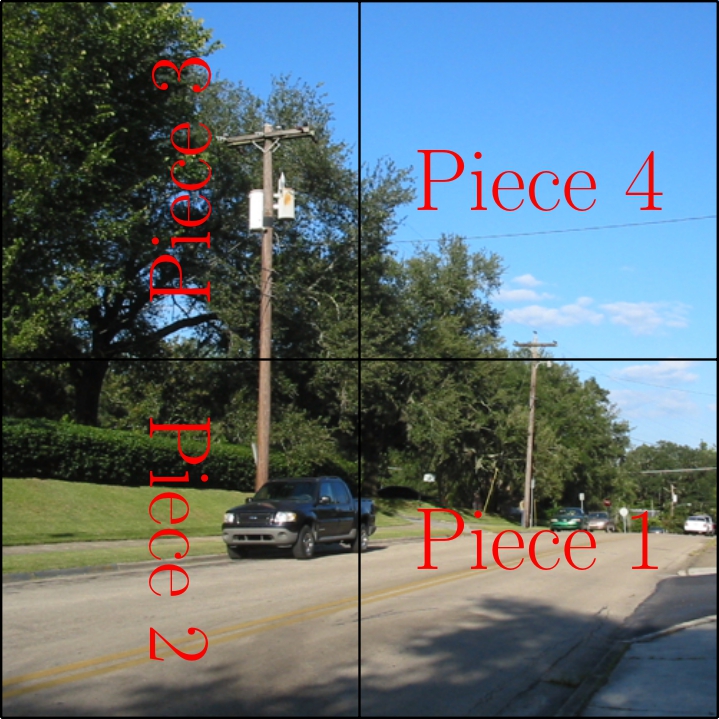}
    \caption{}
    \label{subfig:type_2_labeling:solution}
\end{subfigure}
\par\smallskip
\begin{subfigure}[b]{0.47\textwidth}
    \centering
    \includegraphics[height=21mm]{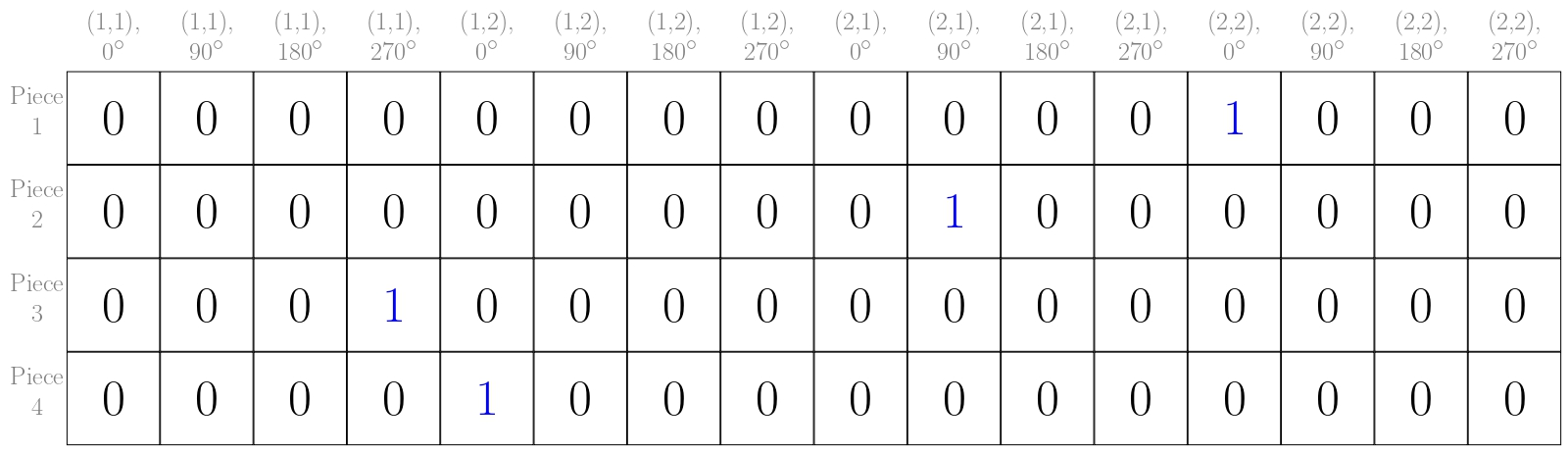}
    \caption{}
    \label{subfig:type_2_labeling:labeling}
\end{subfigure}
\caption{Type 2 puzzle feasible labeling structure. \subref*{subfig:type_2_labeling:puzzle} is a $2 \times 2$ puzzle and \subref*{subfig:type_2_labeling:solution} is its  perfect solution. The corresponding type 2 permutation labeling matrix is shown in \subref*{subfig:type_2_labeling:labeling}.}
\label{fig:type_2_labeling}
\end{figure}

Under this formulation several complications immediately arise. First, the labeling matrix is no longer square as $m = 4n$, and thus the solution \textit{cannot} be a permutation matrix. Second, a feasible Type 2 reconstruction requires to place each piece in \textit{some} orientation at a unique position on the puzzle grid. In fact, in order to represent feasible solutions, the labeling matrix should include a single $1$ in each row (just as in Type 1 puzzles), but at the same time a single $1$ in every group of 4 columns that represent the different orientations of the same position. A labeling for which the matrix representation meets this description is referred to as \textit{type 2 permutation labeling} (see Figure~\ref{fig:type_2_labeling}).
We note that similar to the Type 1 case, in the space of Type 2 feasible solutions the RL optimization seeks to maximize the sum of compatibilities between adjacent pieces.

\begin{figure}
\centering
\begin{subfigure}[b]{0.09\textwidth}
    \centering
    \includegraphics[height=14mm]{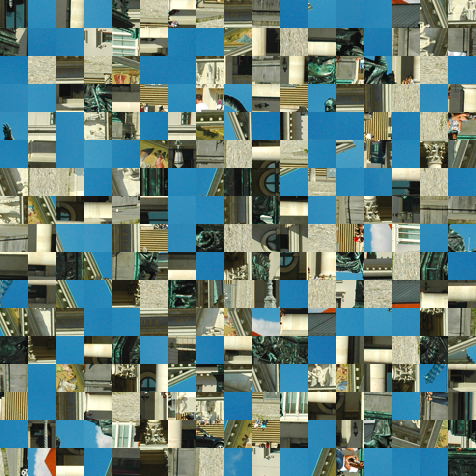}
    \caption{}
    \label{subfig:type_2_possible_reconstructions:square_puzzle}
\end{subfigure}
\begin{subfigure}[b]{0.09\textwidth}
    \centering
    \includegraphics[height=14mm]{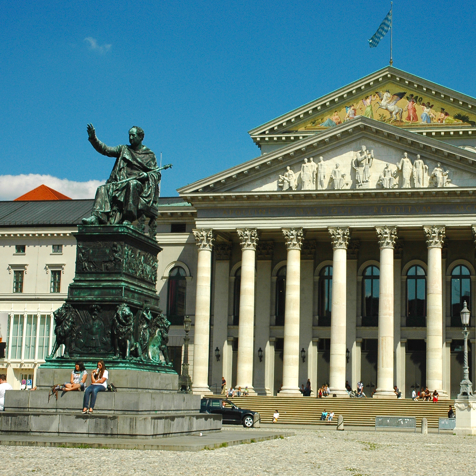}
    \caption{}
    \label{subfig:type_2_possible_reconstructions:square_sol_0}
\end{subfigure}
\begin{subfigure}[b]{0.09\textwidth}
    \centering
    \includegraphics[height=14mm]{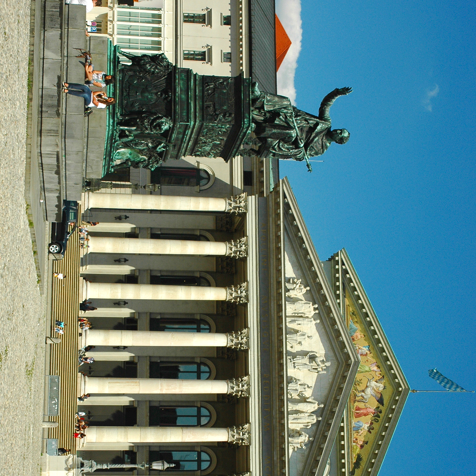}
    \caption{}
    \label{subfig:type_2_possible_reconstructions:square_sol_90}
\end{subfigure}
\begin{subfigure}[b]{0.09\textwidth}
    \centering
    \includegraphics[height=14mm]{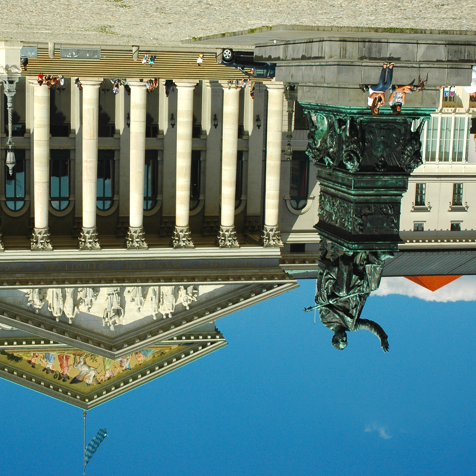}
    \caption{}
    \label{subfig:type_2_possible_reconstructions:square_sol_180}
\end{subfigure}
\begin{subfigure}[b]{0.09\textwidth}
    \centering
    \includegraphics[height=14mm]{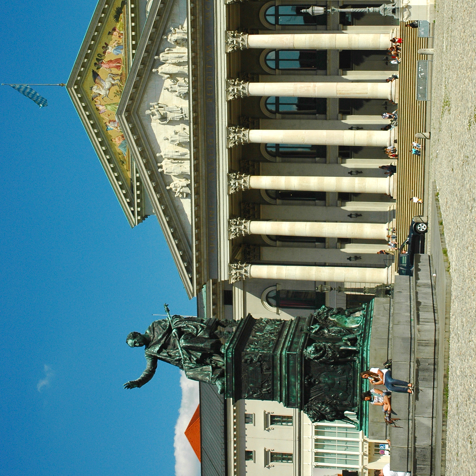}
    \caption{}
    \label{subfig:type_2_possible_reconstructions:square_sol_270}
\end{subfigure}
\par\smallskip
\begin{subfigure}[b]{0.15\textwidth}
    \centering
    \includegraphics[height=15mm]{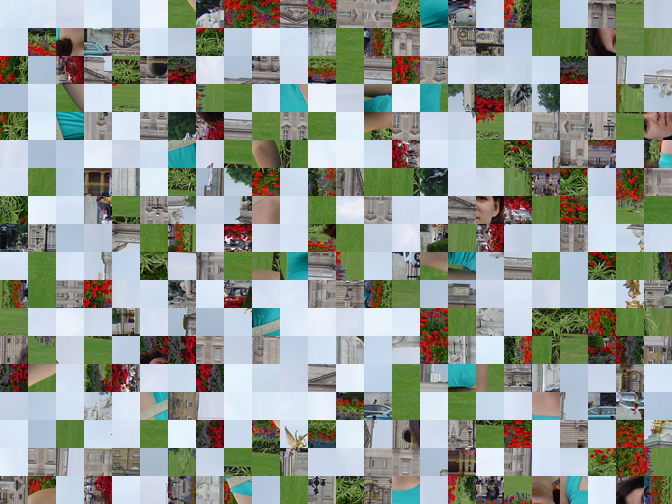}
    \caption{}
    \label{subfig:type_2_possible_reconstructions:rect_puzzle}
\end{subfigure}
\begin{subfigure}[b]{0.15\textwidth}
    \centering
    \includegraphics[height=15mm]{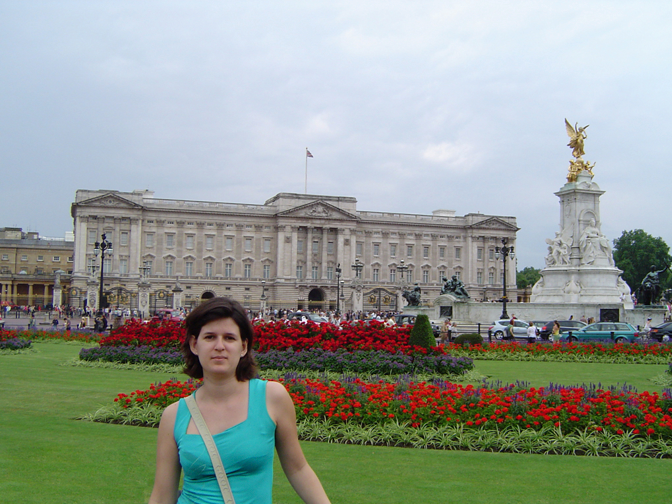}
    \caption{}
    \label{subfig:type_2_possible_reconstructions:rect_sol_0}
\end{subfigure}
\begin{subfigure}[b]{0.15\textwidth}
    \centering
    \includegraphics[height=15mm]{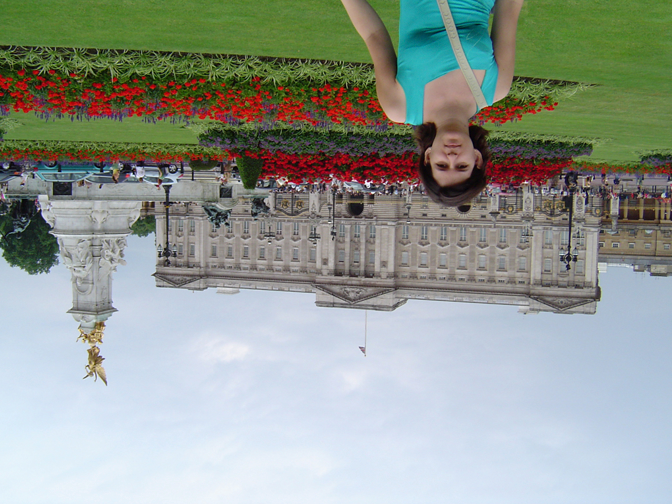}
    \caption{}
    \label{subfig:type_2_possible_reconstructions:rect_sol_180}
\end{subfigure}
\caption{Possible reconstructions for Type 2 puzzles. \subref*{subfig:type_2_possible_reconstructions:square_puzzle} is a square-shaped puzzle, for which \subref*{subfig:type_2_possible_reconstructions:square_sol_0}, \subref*{subfig:type_2_possible_reconstructions:square_sol_90}, \subref*{subfig:type_2_possible_reconstructions:square_sol_180} and \subref*{subfig:type_2_possible_reconstructions:square_sol_270} are four perfect reconstructions.
\subref*{subfig:type_2_possible_reconstructions:rect_puzzle} is a rectangular-shaped puzzle, for which \subref*{subfig:type_2_possible_reconstructions:rect_sol_0} and  \subref*{subfig:type_2_possible_reconstructions:rect_sol_180} are two perfect reconstructions.}
\label{fig:type_2_possible_reconstructions}
\end{figure}

To complete the description of Type 2 problems as RL processes, we need to define the initial labeling. The situation here is again somewhat more complicated since unlike for Type 1 puzzles, that have a single perfect reconstruction, any Type 2 puzzle has at least 2 perfect reconstructions due to global image rotation (see Figure~\ref{fig:type_2_possible_reconstructions}). In order to encourage convergence to one of these solutions, the dynamics should be biased towards a specific solution rather than \enquote{favoring} all solutions equally.
To address this, we set the initial labeling as the barycenter of the search space, while limiting some piece $b_{i_1}$ to some
orientation $\theta_1$:
\begin{equation}\label{eq:type_2_initial_labeling}
\small
p_{i}(\lambda) = p_{i}(((x_\lambda, y_\lambda), \theta_{\lambda})) =
\begin{cases*}
  \frac{1}{m}, & if $i \neq i_1$,\\
  \frac{1}{n}, & if $i=i_1, \theta_{\lambda}=\theta_1$,\\
  0, & if $i=i_1, \theta_{\lambda} \neq \theta_1$\,.
\end{cases*}
\end{equation}
When puzzles are square in shape (i.e., $N=M$), the orientation $\theta_1$ is chosen randomly among the 4 possible ones and serves as an ``angular anchor'' that pulls the entire solution towards the proper global orientation.
Doing the same for rectangular $N \times M$ puzzles is problematic because the selected orientation of piece $b_{i_1}$, and thus the preferred global orientation of the solution, might conflict the rectangular dimensions of the puzzles. To cope with this problem we run the algorithm twice, first with an initial state derived from a randomly selected orientation $\theta_1$, and once again with an initial state derived from $\theta_1+90\degree$. This ensures that one of the computations could converge to a correct reconstruction, which is identified and selected by its larger ALC. 
Finally, in order to maximize the influence of the bias obtained from the initial state, piece $b_{i_1}$ is heuristically selected as the one that maximizes the sum of potential compatibilities in its 4-neighborhood.

\subsubsection{Type 2 Solver}

\begin{figure}
\centering
\begin{subfigure}[b]{0.47\textwidth}
    \centering
    \includegraphics[height=22.5mm]{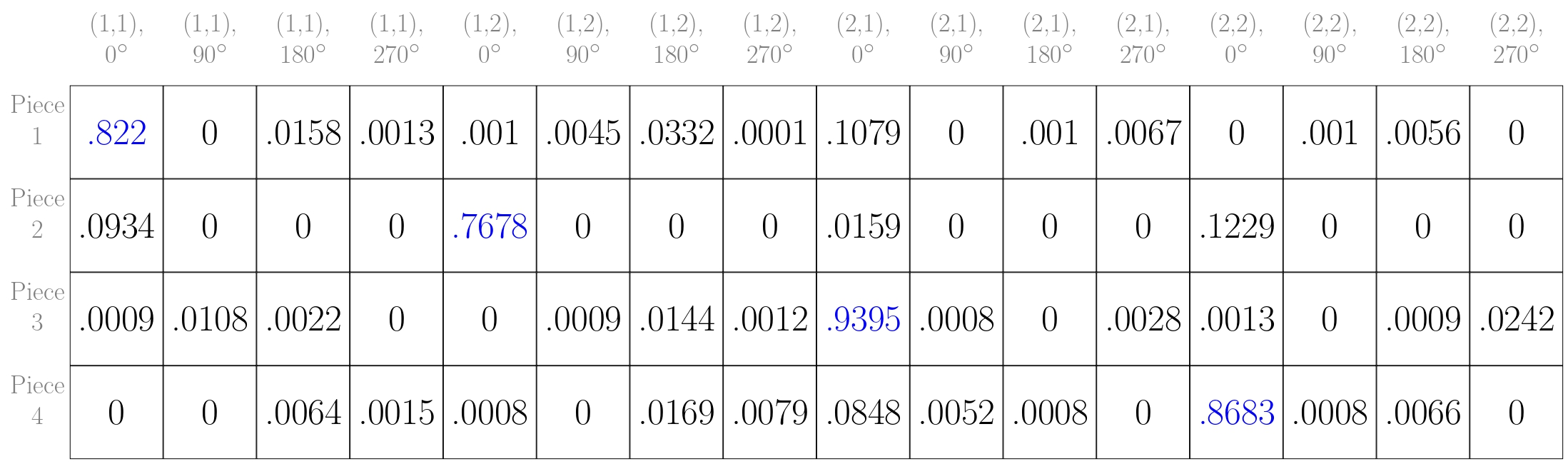}
\end{subfigure}
\par\smallskip
\begin{subfigure}[b]{0.47\textwidth}
    \centering
    \includegraphics[height=22.5mm]{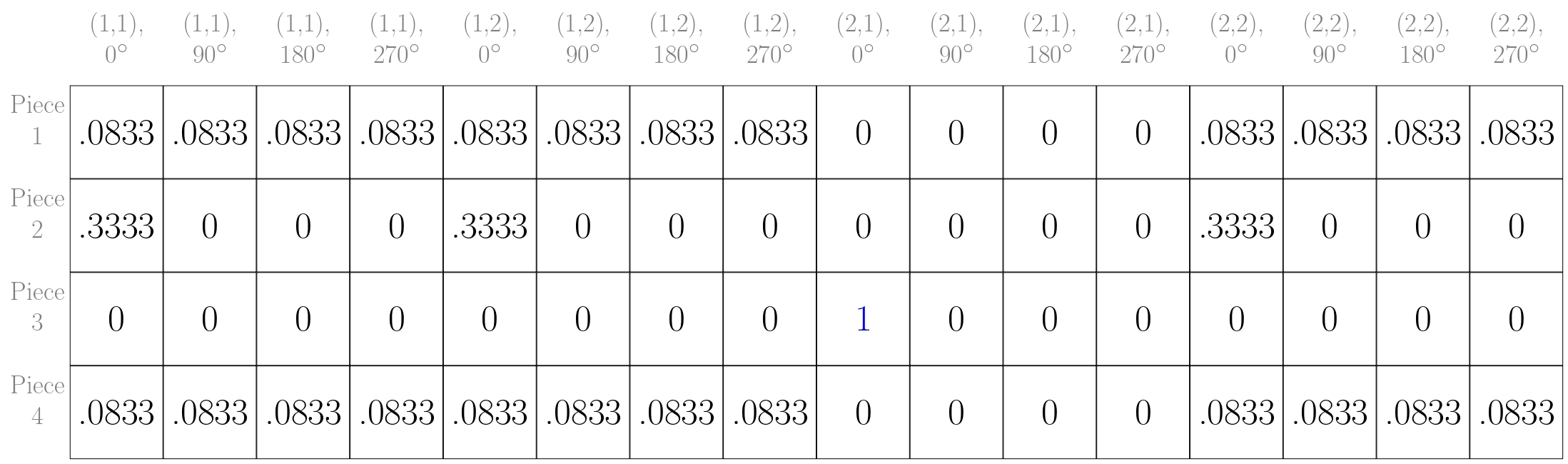}
\end{subfigure}
\caption{
The first phase of our Type 2 solver on a $2 \times 2$ puzzle, with $\alpha = 0.7$. The RL phase begins from the initial labeling and terminates once labeling values cross $\alpha$ (top matrix, obtained after $3$ RL iterations). The best supra-threshold entry at this point is $p_3(((2,1), 0\degree))$. After anchoring piece $b_3$ to position $(2,1)$ at orientation $0\degree$ we obtain the bottom matrix. Note that $b_2$ is allowed only at orientation $0\degree$ since the initial labeling was set this way.}
\label{fig:type_2_solver_phase}
\end{figure}

Not unlike the Type 1 solver that directs the process to a permutation labeling, the Type 2 solver needs to lead the process to a type 2 permutation labeling. However, to \enquote{anchor} an entry $p_i(((x_\lambda, y_\lambda), \theta_\lambda))$ requires to revise the intervention at the end of each RL phase to manipulate the labeling in a slightly different way:
\begin{enumerate}[(1), noitemsep]
\item Anchoring piece $b_i$ to position $(x_\lambda, y_\lambda)$ in orientation $\theta_\lambda$ and preventing its placement at any other position (regardless of orientation). Formally we set $p_i(((x_\lambda, y_\lambda), \theta_\lambda))~=~1$ and $p_i(\mu)~=~0,\; \forall~\mu~\neq~((x_\lambda, y_\lambda), \theta_\lambda)$. 
\item Preventing a future placement of other pieces at position $(x_\lambda, y_\lambda)$ regardless of their orientation. This is done by setting $p_j(((x_\lambda, y_\lambda), \theta))~=~0,\; \forall~\theta,\;~j~\neq~i$.
\item Resetting entries $p_j(((x_\mu, y_\mu), \theta_\mu))$ that involve pieces $b_j$ and positions $(x_\mu, y_\mu)$ for which anchoring is yet made. This is done by setting these entries to the reciprocal of the number of combinations of unassigned positions and possible orientations (i.e., the number of unassigned positions multiplied by $4$). If piece $b_{i_1}$ (used for setting the initial labeling) is not anchored, we reset its values so only orientation $\theta_1$ is allowed.
\end{enumerate}
Figure~\ref{fig:type_2_solver_phase} demonstrates the effect of these 3 steps after the completion of the first RL phase on a $2 \times 2$ Type 2 puzzle. The shifting of the anchored block at the end of each phase is handled as before.

\subsection{Piece Compatibility Measure}\label{sec:piece_compatibility}

At the foundations of any puzzle solver is some measure of affinity between puzzle pieces that allows to quantify the quality of immediate neighborhood relationships. In our algorithm it is integrated in RL compatibility coefficients.

As mentioned earlier, piece compatibility measures are conventionally based on a more basic measure of piece dissimilarity. 
While various dissimilarity measures have been proposed, we use the Mahalanobis gradient dissimilarity proposed by Son et~al.~(2018). This dissimilarity measure evaluates the disagreement of placing two adjacent pieces by computing the Mahalanobis distances between the actual gradients and expected gradients with respect to the distribution of predicted gradients, where the gradients considered are both across and along the adjoining boundaries of pieces. Please refer to Son et~al.~(2018) for a full description.

Given the dissimilarity, it is possible to convert it to a compatibility function in various ways, the most immediate of which is simply negating the dissimilarities. Here we take a more elaborate approach that is based on the observation that our solver, and RL processes in general, perform better when the compatibility values are well spaced and the difference between compatible and incompatible values is significant. This was already observed in other solvers, and here we follow Andal{\'o} et~al.~(2016) who improved the basic idea from Pomeranz et~al.~(2011) and defined
\begin{equation}\label{eq:andalo_comp_function}
\small
C(b_i, b_j, R) = exp\bigg(-\varphi_{i,R}(j) -\frac{D(b_i, b_j, R)}{quartile(i,R)}\bigg)\,,
\end{equation}
where $quartile(i,R)$ is the quartile dissimilarity among all dissimilarities in relation $R$ to piece $b_i$, 
and $\varphi_{i,R}(j)$ is the rank of the dissimilarity $D(b_i, b_j, R)$ in the increasingly ordered dissimilarities in relation $R$ to piece $b_i$. The $quartile(i, R)$ term is included to consider the scattering of dissimilarity values (and not just the values themselves), and the $\varphi_{i,R}(j)$ rank term results in \enquote{spacing} the computed compatibility values.

Inspired by such ideas, we propose a new compatibility measure:
\begin{equation}\label{eq:second_comp_function}
\small
\begin{split}
& C(b_i, b_j, R)=\\
& \begin{cases*}
  1, & if $D(b_i, b_j, R) = p_{avg}(i,R;k) = 0$,\\
  \big(1-\frac{D(b_i, b_j, R)}{p_{avg}(i,R;k)}\big)^{\varphi_{i,R}(j)}, & if $D(b_i, b_j, R) \leq p_{avg}(i,R;k)$\\
  & and $p_{avg}(i,R;k) > 0$,\\ 
  0, & otherwise\,,\\
\end{cases*}
\end{split}
\end{equation}
where $p_{avg}(i,R;k)$ is the average dissimilarity from the first value till the $k$ percentile among all dissimilarities in relation $R$ to piece $b_i$.

This measure offers two advantages over the measure from Eq.~\ref{eq:andalo_comp_function}: (1) dissimilarities scattering information is determined by multiple values rather than one value (the quartile dissimilarity). Moreover, we introduce $k$ as a parameter to avoid limitation to $k=25$ (the quartile). (2) It better balances the $D(b_i, b_j, R)$ and $\varphi_{i,R}(j)$ terms since high $\varphi_{i,R}(j)$ does not necessarily guarantee low compatibility score.

An important challenge that most global optimization solvers face is due to \textit{constant pieces}, or puzzle pieces that have the same constant color value throughout, which happens mostly in sky regions that turned white due to pixel saturation. Constant pieces maintain maximal compatibility with other constant pieces (and possibly with more pieces), thus it becomes difficult for the RL process to converge to a labeling that unambiguously determines their positions.
To alleviate this, we slightly adjust the compatibility for puzzles with more than two constant pieces:
\begin{equation}\label{eq:constant_pieces_manipulation}
\small
C'(b_i, b_j, R)=
\begin{cases*}
    max\{0, X\}, & if there exists constant \\
    & piece $b_k$ s.t. $C(b_i, b_k, R)=1$ \\
    & and $C(b_k, b_j, R)=1$,\\
    C(b_i, b_j, R), & otherwise\,, \\
\end{cases*}
\end{equation}
where $X\sim U(-4,1)$ is a random variable.
This adjustment re-draws all maximal compatibilities for constant pieces (and after rectification set it to $0$ with probability of $0.8$), which sparsifies the compatibility coefficients and assists convergence.

\section{\uppercase{Experimental results}}

To test the suggested puzzle solvers, we used the $20$ puzzles of $432$ pieces from the MIT data set proposed by Cho et~al.~(2010) and the $20$ puzzles of $540$ pieces from the McGill data set proposed by Pomeranz et~al.~(2011). All puzzles contain pieces of size of $28 \times 28$ pixels. We evaluated performance with the following three performance measures:
(1) \textit{Direct comparison} (DC; Cho et~al., 2010) is the fraction of pieces that are in the same position (and orientation for Type 2 puzzles) in the assembly solution and the ground truth solution. 
(2) \textit{Neighbor comparison} (NC; Cho et~al., 2010) is the fraction of pairwise piece matchings that exist in the ground truth solution. 
(3) \textit{Perfect reconstruction} (PR; Gallagher, 2012) is a binary indication of whether the assembly solution is identical to the ground truth. For an entire data set, PR becomes the number of perfectly solved puzzles.

\begin{table}[t]
\centering
\caption{Comparison of Type 1 solvers.}
\resizebox{!}{14mm}{
\begin{tabular}{ |c|c|c|c|c|c|c| } 
    \hline
    \multirow{2}{*}{Method} & \multicolumn{3}{|c|}{The MIT data set} & \multicolumn{3}{|c|}{The McGill data set}\\ 
    \cline{2-7}
    & DC & NC & PR & DC & NC & PR\\ 
    \hline
    Andal{\'o} et~al.~(2016) & {96\%} & {95.6\%} & {13} & {90.8\%} & {95.3\%} & {13}\\
    \hline
    Gallagher~(2012) & {95.3\%} & {95.1\%} & {12} & {-} & {-} & {-}\\ 
    \hline
    Paikin and Tal~(2015) & {96.2\%} & {95.8\%} & {13} & {93.2\%} & {96.1\%} & {13}\\ 
    \hline
    Pomeranz et~al.~(2011) & {94\%} & {95\%} & {13} & {83.5\%} & {90.9\%} & {9}\\ 
    \hline
    Sholomon et~al.~(2013) & {86.2\%} & {96.2\%} & {9} & {92.8\%} & {96\%} & {8}\\
    \hline
    Son et~al.~(2018) & {95.8\%} & {95.6\%} & {-} & {95.4\%} & {97\%} & {-}\\ 
    \hline
    Khoroshiltseva et~al.~(2021) & {79.1\%} & {87.9\%} & {6} & {25.6\%} & {64.5\%} & {0}\\ 
    \hline
    Ours & {95.5\%} & {95.4\%} & {13} & {62.9\%} & {87.8\%} & {6}\\ 
    \hline
\end{tabular}
}
\label{table:type_1_results}
\end{table}

\begin{table}[t]
\centering
\caption{Comparison of Type 2 solvers.}
\resizebox{!}{12mm}{
\begin{tabular}{ |c|c|c|c|c|c|c| } 
    \hline
    \multirow{2}{*}{Method} & \multicolumn{3}{|c|}{The MIT data set} & \multicolumn{3}{|c|}{The McGill data set}\\ 
    \cline{2-7}
    & DC & NC & PR & DC & NC & PR\\ 
    \hline
    Gallagher~(2012) & $82.2\%$ & $90.4\%$ & 9 & {-} & {-} & {-}\\ 
    \hline
    Huroyan et~al.~(2020) & $94.8\%$ & $95.2\%$ & 13 & {88.3\%} & {92.2\%} & {13}\\ 
    \hline
    Paikin and Tal~(2015) & $95.4\%$ & $95.4\%$ & - & {-} & {-} & {-}\\ 
    \hline
    Sholomon et~al.~(2014) & $94.6\%$ & $94.9\%$ & 10 & {89.6\%} & {92\%} & {8}\\ 
    \hline
    Son et~al.~(2018) & $95.8\%$ & $95.6\%$ & 12 & {92.3\%} & {94.5\%} & {11}\\ 
    \hline
    Ours & {87.1\%} & {93.9\%} & 5 & {25.8\%} & {78\%} & {0}\\ 
    \hline
\end{tabular}}
\label{table:type_2_results}
\end{table}

\nocite{paikin2015solving}

The only random part in our solvers is the constant pieces compatibility adjustment (Eq.~\ref{eq:constant_pieces_manipulation}). Otherwise, solvers are fully deterministic for a given input. For this reason, the reported performance for puzzles with constant pieces is the average of $10$ runs while the rest of the puzzles were solved once. In all experiments, the RL convergence threshold $\epsilon$ was set to $10^{-4}$, while the $\alpha$ anchoring threshold was set to $0.7$. 
Dissimilarity computations are based on the CIELAB color space. Compatibility computations (Eq.~\ref{eq:second_comp_function}) use $k=3$ and $k=1.5$ for Type 1 and 2 solvers, respectively.

\begin{figure}
\centering

\begin{subfigure}[b]{0.47\textwidth}
    \centering
    \includegraphics[height=18.05mm]{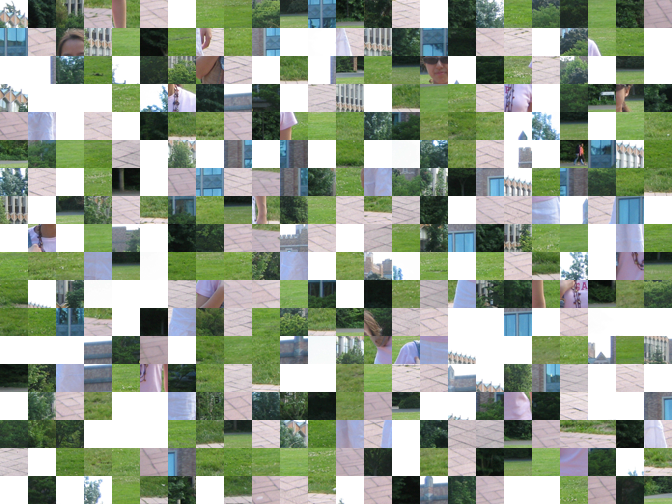}
    \includegraphics[height=18.05mm]{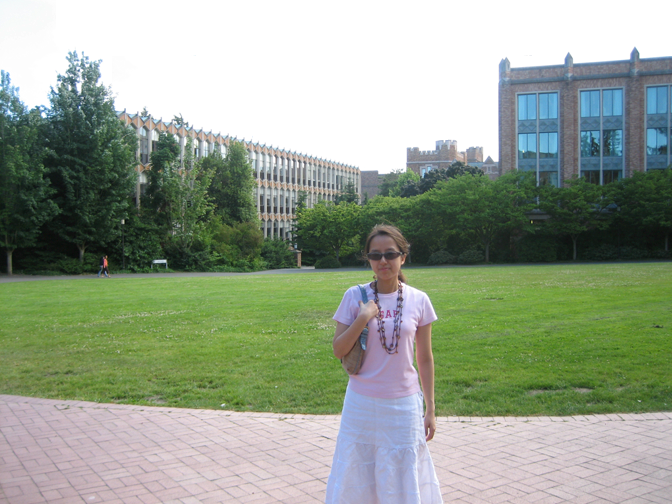}
    \includegraphics[height=18.05mm]{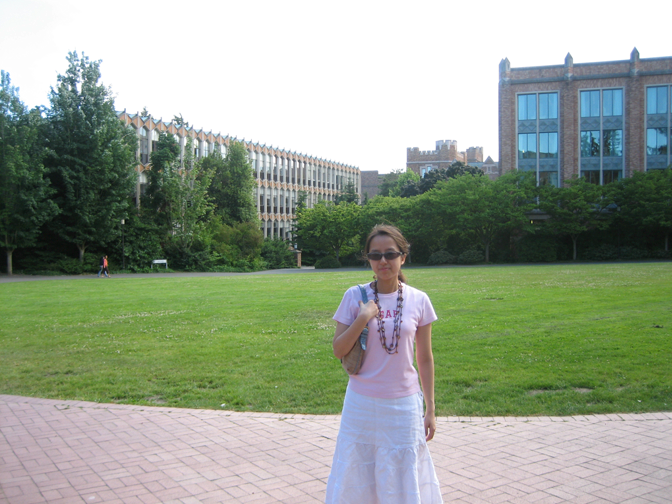}
    \caption{}
    \label{subfig:results:type1_constant}
\end{subfigure}
\par\smallskip
\begin{subfigure}[b]{0.47\textwidth}
    \centering
    \includegraphics[height=17.85mm]{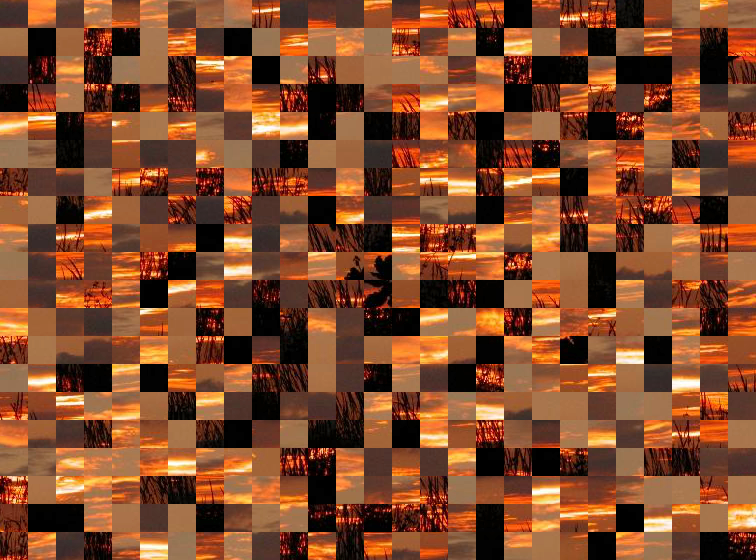}
    \includegraphics[height=17.85mm]{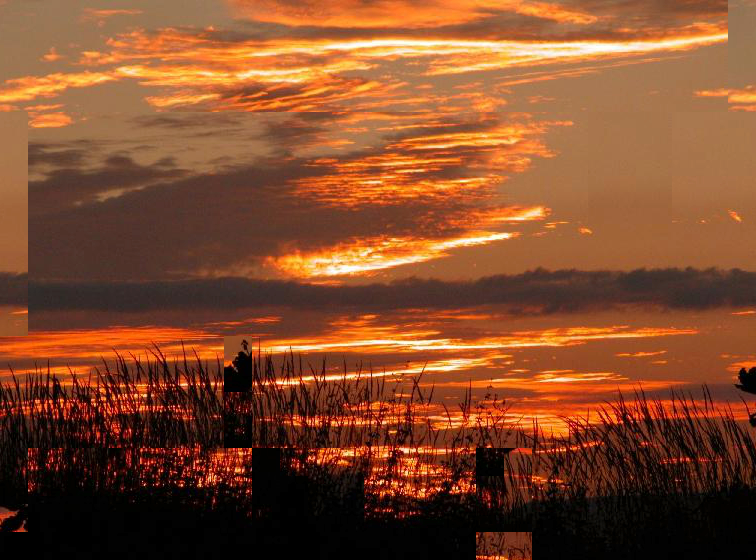}
    \includegraphics[height=17.85mm]{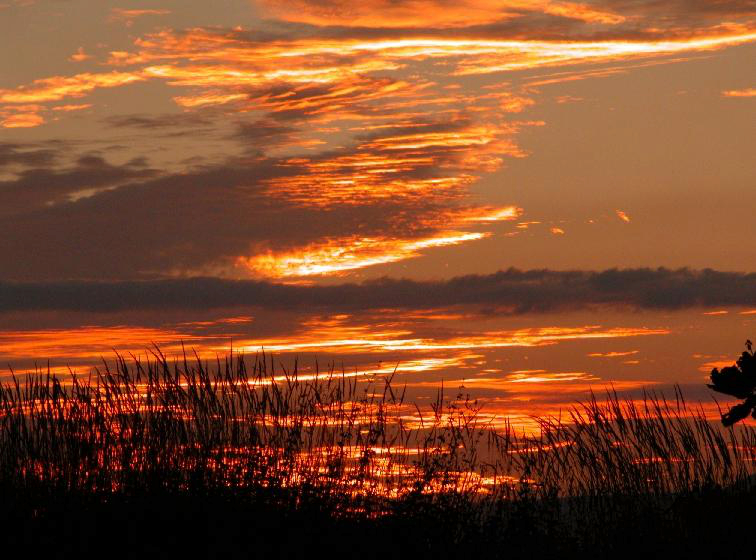}
    \caption{}
    \label{subfig:results:type1_offset}
\end{subfigure}
\par\smallskip
\begin{subfigure}[b]{0.47\textwidth}
    \centering
    \includegraphics[height=18.05mm]{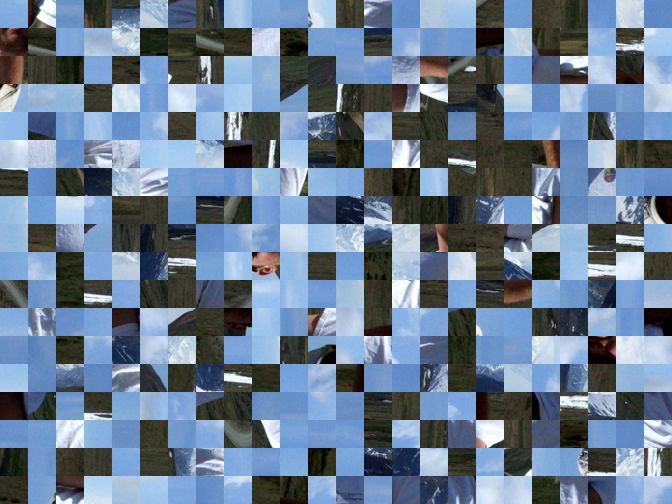}
    \includegraphics[height=18.05mm]{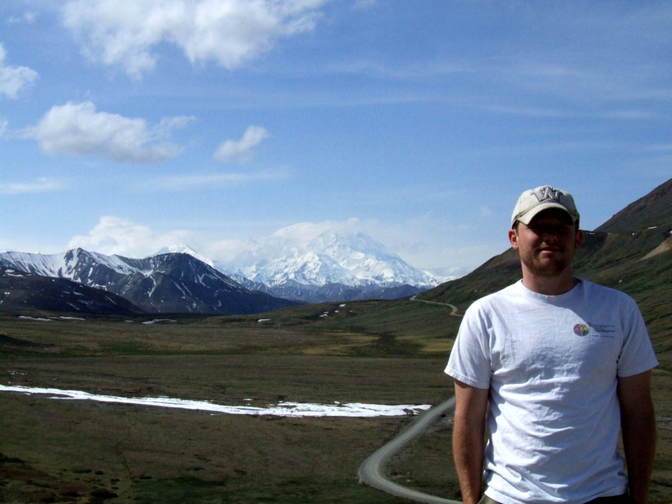}
    \includegraphics[height=18.05mm]{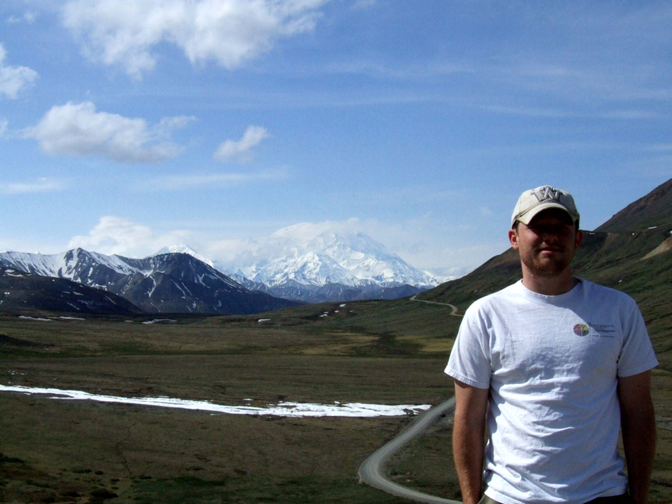}
    \caption{}
    \label{subfig:results:type2_perfect}
\end{subfigure}
\par\smallskip
\begin{subfigure}[b]{0.47\textwidth}
    \centering
    \includegraphics[height=17.85mm]{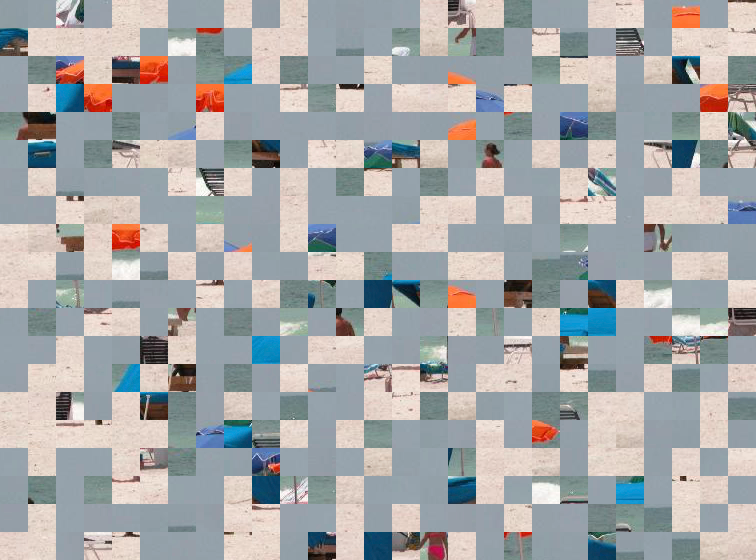}
    \includegraphics[height=17.85mm]{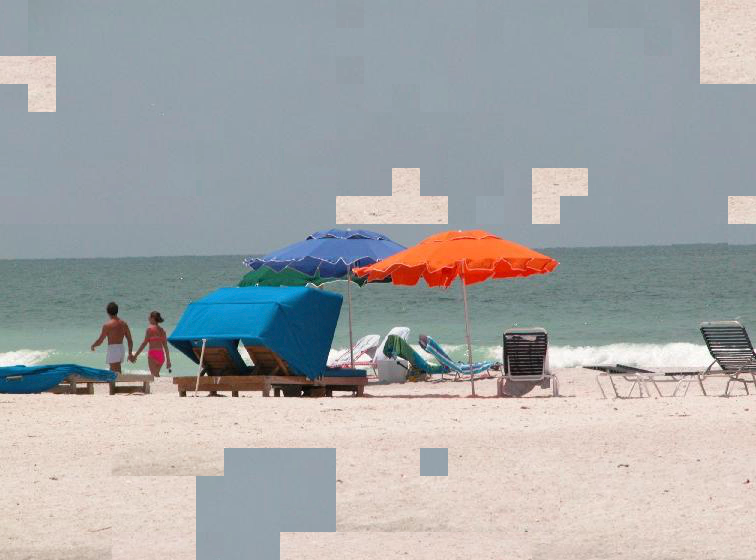}
    \includegraphics[height=17.85mm]{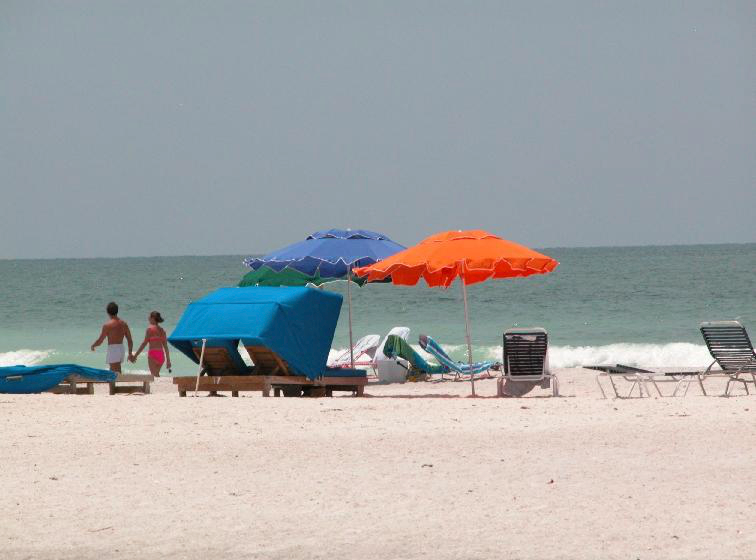}
    \caption{}
    \label{subfig:results:type1_bad}
\end{subfigure}
\par\smallskip
\begin{subfigure}[b]{0.47\textwidth}
    \centering
    \includegraphics[height=18.05mm]{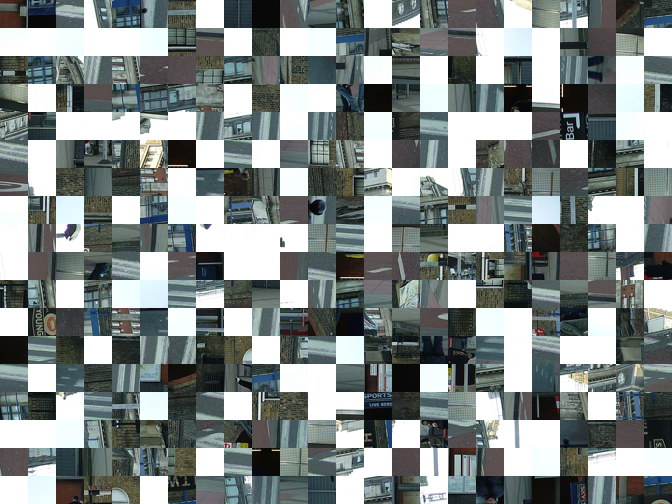}
    \includegraphics[height=18.05mm]{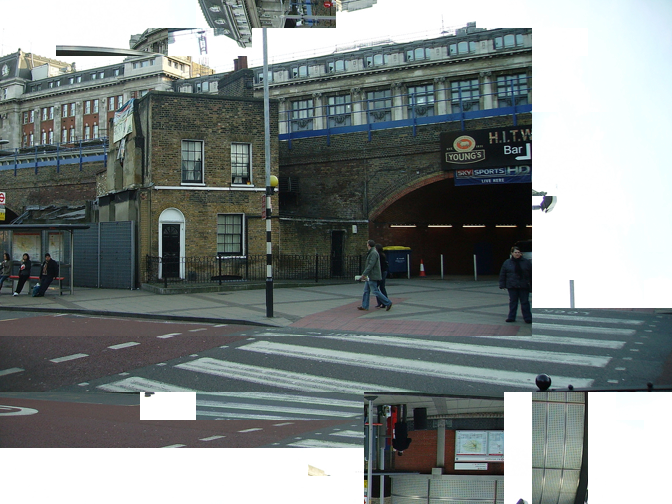}
    \includegraphics[height=18.05mm]{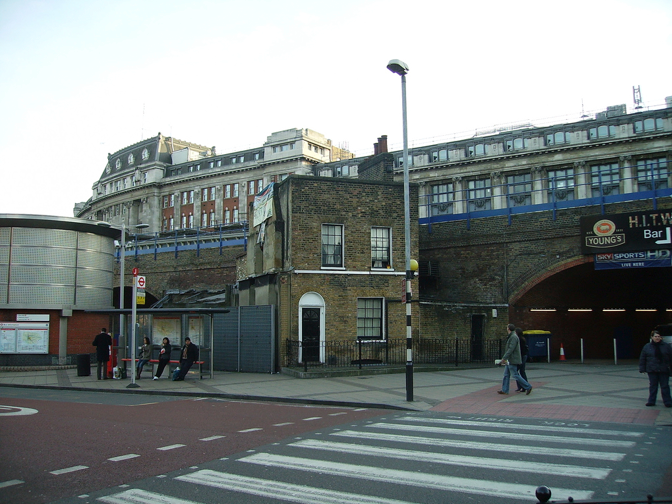}
    \caption{}
    \label{subfig:results:type2_bad}
\end{subfigure}
\caption{Input puzzle, RL solver result, and ground truth image are shown left to right. \subref*{subfig:results:type1_constant} shows a typical result on a Type 1 puzzle with constant pieces (sky pieces are constant white). While the solution is visually perfect, its DC and NC scores are relatively low ($83.8\%$ and $81.8\%$, respectively) due to wrong internal ordering of constant pieces. \subref*{subfig:results:type1_offset} shows a Type 1 solution with a very low DC score ($6.3\%$) even though the solution is visually excellent, due to offset of one column. 
\subref*{subfig:results:type2_perfect} shows a perfectly solved Type 2 puzzle. 
\subref*{subfig:results:type1_bad} and \subref*{subfig:results:type2_bad} exemplify an occasional failure of the Type 1 and 2 solvers (respectively) in which sub-blocks of pieces are assembled correctly, but in wrong position (and rotation for Type 2) relative to each other.}
\label{fig:results}
\end{figure}

Table~\ref{table:type_1_results} presents the Type~1 solver performance. Our main goal in this paper is a progression in performance of RL solvers and indeed the multi-phase method performs better than Khoroshiltseva et~al.~(2021). Equally important, results are highly competitive with many prior methods on the MIT data set and approach the MIT data set upper bounds (DC, NC, and PR upper bounds are $96.7\%$, $96.4\%$ and $15$, respectively; Son et~al., 2014), which are below perfect since no solver is expected to internally order identical constant pieces correctly. 
The results of prior methods were taken from their papers, while the results of Khoroshiltseva et~al.~(2021) were obtained by our implementation for their solver with Sinkhorn-Knopp balancing.
Table~\ref{table:type_2_results} presents our Type 2 results compared to other methods. Because Type 2 solutions may be rotated compared to the ground truth (see Figure~\ref{fig:type_2_possible_reconstructions}), we evaluate only the rotation that achieves the highest DC. Figure~\ref{fig:results} shows Type 1 and 2 results.

\section{\uppercase{Conclusion}}

This paper suggests a step forward in RL puzzle solving based on a novel multi-phase RL optimization approach that is fully automatic, uses no prior information, guarantees feasible solutions, and is applicable for both Type 1 and 2 puzzles. 
Future work should handle failure cases (see Figures~\ref{subfig:results:type1_bad}-\ref{subfig:results:type2_bad}), which could, we believe, be alleviated within the RL framework. For example, by identifying correctly assembled block (such as done by Pomeranz et~al., 2011), modifying the compatibility coefficients accordingly and running the RL process again.

We believe that the proposed multi-stage approach could be used as a general RL-based scheme for other hard optimization problems (not necessarily puzzles) for solving permutation challenges (cf. Type 1 cases) or even more complicated permutation constraints (cf. Type 2 cases). For example, it is likely that the multi-phase approach could be applied to much larger scale instances of the traveling salesman problem than previously attempted with RL~(Pelillo, 1993).

\nocite{pelillo1993relaxation}

\section*{\uppercase{Acknowledgements}}

This project has received funding from the European Union’s Horizon 2020 research and innovation programme under grant agreement No 964854, the Helmsley Charitable Trust through the ABC Robotics Initiative, and the Frankel Fund of the Computer Science Department at Ben-Gurion University.

\bibliographystyle{apalike}
{\small
\bibliography{paperbib}}

\end{document}